\documentclass[conference]{IEEEtran}
\IEEEoverridecommandlockouts
\usepackage{url}
\usepackage{bm}
\usepackage{color}
\usepackage{fancyhdr} 
\usepackage{amsmath}
\usepackage{array}
\usepackage{graphicx}
\usepackage{booktabs}
\usepackage{makecell}
\usepackage{comment}
\usepackage{hyperref}
\usepackage{tikz}
\usepackage{textcomp}
\usepackage{lipsum}
\usepackage{setspace}
\usepackage{multirow}
\usepackage{amsfonts,amssymb}
\usepackage[ruled,vlined]{algorithm2e}
\usepackage[explicit]{titlesec}
\usepackage{cite}
\usepackage{amsmath,amssymb,amsfonts}
\usepackage{algorithmic}
\usepackage{graphicx}
\usepackage{textcomp}
\usepackage{xcolor}
\def\BibTeX{{\rm B\kern-.05em{\sc i\kern-.025em b}\kern-.08em
    T\kern-.1667em\lower.7ex\hbox{E}\kern-.125emX}}
\makeatletter
\renewcommand\footnoterule{
 \kern-3\p@
 \hrule\@width1\columnwidth
 \kern2.6\p@
}
\makeatletter
\begin{document}
\title{An Approach for Multi-Object Tracking with Two-Stage Min-Cost Flow\\
\thanks{*Corresponding author.This work was supported by Beijing Municipal Natural Science Foundation No. 4234085}
}
\author{\IEEEauthorblockN{1\textsuperscript{st} Huining Li}
\IEEEauthorblockA{\textit{School of Electronic and Information} \\
\textit{Beihang University}\\
Beijing, China \\
lhn2202520@buaa.edu.cn}
\and
\IEEEauthorblockN{2\textsuperscript{nd} Yalong Jiang$^{*}$}
\IEEEauthorblockA{\textit{Unmanned System Research Institute} \\
\textit{Beihang University}\\
Beijing, China \\
AllenYLJiang@outlook.com}
\and
\IEEEauthorblockN{3\textsuperscript{rd} Xianlin Zeng}
\IEEEauthorblockA{\textit{School of Electronic and Information} \\
\textit{Beihang University}\\
Beijing, China \\
zengxianlin@buaa.edu.cn}
\and 
\IEEEauthorblockN{4\textsuperscript{th} Feng Li}
\IEEEauthorblockA{\textit{China Academy of Information and Communications Technology} \\
Beijing, China \\
lifeng@caict.ac.cn}
\and
\IEEEauthorblockN{5\textsuperscript{th} Zhipeng Wang}
\IEEEauthorblockA{\textit{School of Electronic and Information} \\
\textit{Beihang University}\\
Beijing, China \\
wangzhipeng@buaa.edu.cn}
}
\maketitle
\thispagestyle{fancy}
\lhead{}
\rhead{}
\renewcommand{\headrulewidth}{0.0pt}
\lfoot{\small{979-8-3503-9416-0/23/\$31.00 ©2023 IEEE} }
\cfoot{}
\rfoot{}
\begin{abstract}
The minimum network flow algorithm is widely used in multi-target tracking.  However, the majority of the present methods concentrate exclusively on
minimizing cost functions whose values may not indicate accurate solutions under occlusions. In this paper, by exploiting the properties of tracklets intersections and low-confidence detections, we develop a two-stage tracking pipeline with an intersection mask that can accurately locate inaccurate tracklets which are corrected in the second stage. 
Specifically, we employ the minimum network flow algorithm with high-confidence detections as input in the first stage to obtain the candidate tracklets that need correction. Then we leverage the intersection mask to accurately locate the inaccurate parts of candidate tracklets. The second stage utilizes low-confidence detections that may be attributed to occlusions for correcting inaccurate tracklets. This process constructs a graph of nodes in inaccurate tracklets and low-confidence nodes and uses it for the second round of minimum network flow calculation. 
We perform sufficient experiments on popular MOT benchmark datasets and achieve 78.4 MOTA on the test set of MOT16, 79.2 on MOT17, and 76.4 on MOT20, which shows that the proposed method is effective.
\end{abstract}


\section{Introduction}
Multi-object Tracking (MOT) involves the prediction of the object instances trajectories within a video sequence. This task is fraught with challenges arising from occlusions, objects with unpredictable movement patterns, and objects with similar appearances. Despite these challenges, MOT serves as a vital component in various applications, including action recognition, surveillance, and autonomous driving.
The tracking-by-detection\cite{zhang_bytetrack_2022,bewleySimpleOnlineRealtime2016,milanContinuousEnergyMinimization2014,posseggerOcclusionGeodesicsOnline2014,zhouTrackingObjectsPoints2020,zhangFairMOTFairnessDetection2021,meinhardtTrackFormerMultiObjectTracking,sunTransTrackMultipleObject2021,stadlerModellingAmbiguousAssignments2022,caoObservationCentricSORTRethinking2022} technique is the most commonly used method for Multi-object Tracking (MOT). In this method, object detectors like \cite{geYOLOXExceedingYOLO2021,renFasterRCNNRealTime2015,felzenszwalb2008discriminatively,yang2016exploit} initially identify potential object locations by generating bounding boxes. Subsequently, the challenge of multi-object tracking is transformed into a data association problem, where the goal is to assign these bounding boxes to trajectories that represent the movement paths of separate object instances across time.

Min-cost flow is a common method to solve the association problem in MOT \cite{lizhangGlobalDataAssociation2008,leal-taixeEverybodyNeedsSomebody2011,wang_mussp_2019,lenzFollowMeEfficientOnline2014, liLearningGlobalObjective2022b}. Linear programming (LP) can be utilized to resolve all these association challenges. The cost, which measures the similarity between nodes, determines the association performance. To address this, 
\cite{chari2015pairwise} adds pairwise costs that penalize the overlap between different tracks to the min-cost flow framework.
In \cite{schulterDeepNetworkFlow2017} a differentiable function is formulated to achieve the optimal solution to a smoothed network flow problem. Pairwise association costs are learned through back propagation to capture features. In comparison to \cite{schulterDeepNetworkFlow2017}, \cite{liLearningGlobalObjective2022b} combines a more robust motion observation model with the optimised cost function for data association. 
However, the trajectories obtained by minimizing the total loss are ambiguous. The previous works neglect the properties of inaccurate tracklets. In addition, because low-confidence detection boxes often contain significant noise interference, they do not solve the problem that the minimum network flow is sensitive to false-positive detections.

In this paper, we address incorrect tracklet formation in min-cost flow caused by occlusion. To mitigate the problem and increase robustness against false positives, we propose a two-stage MOT tracker named TSMCF. Firstly we construct a graph using high-confidence detections and use the intersection mask to locate the wrong tracklets. Subsequently, we add low-confidence detections to correct the wrong tracklets within the second graph. Finally, we associate the tracklets to obtain the final results. 

\section{Methods}
In this section, we introduce our approach to multi-object tracking, the method has two stages each of which is implemented with min-cost flow algorithm.
\begin{figure*}[!ht]
    \centering
    \begin{center}
    \includegraphics[width=0.9 \textwidth]{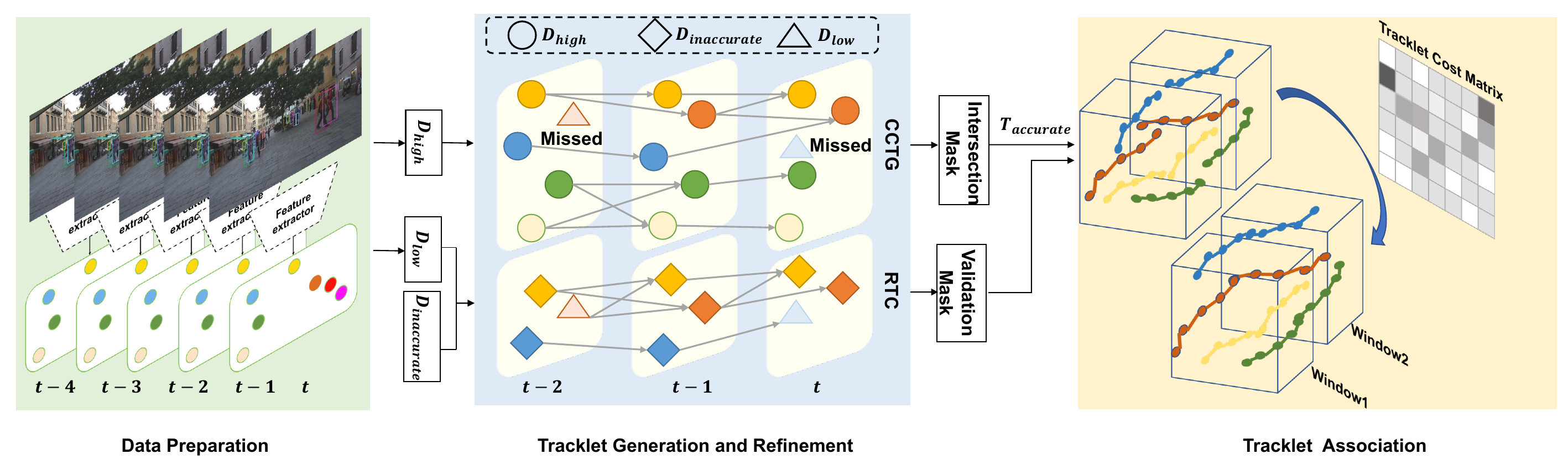}
    \end{center}
    \caption{Outline of the proposed TSMCF tracker. The left portion depicts the data preparation stage. The middle is the key part of the proposed two-stage min-cost flow network for associations within the tracklet, which consists of Coarse Candidate Tracklet Generation (CCTG) and Refined Tracklet Correction (RTC). The right part is the final association process between tracklets. } 
    \label{fig:fig1}
\end{figure*}
\subsection{Minimum-Cost Network Flow Problem}
\begin{figure}[!t]
   \centering
	\normalsize
	\renewcommand{\tabcolsep}{1.3pt} 
	\renewcommand{\arraystretch}{1.3} 
	\includegraphics[width=0.8\columnwidth]{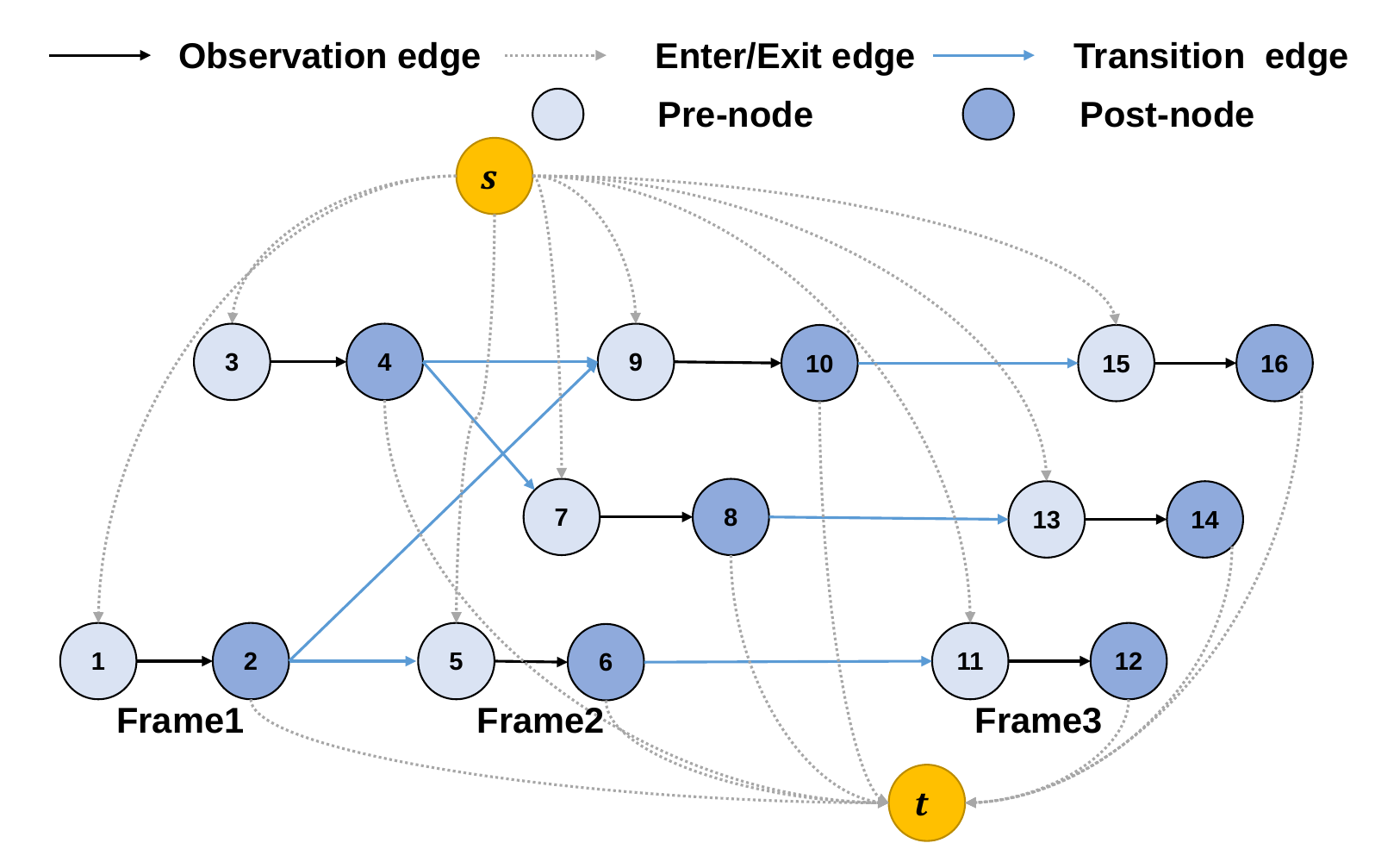}
   \caption{An example of a cost-flow network consists of 8 observations in 3 frames.}
 \label{fig:ssp}
\end{figure}

Considering a collection of detections $\emph{D}=\left \{d_1,...,d_{N_d} \right\}$, where each detection is characterised by its bounding box position $\left(x_l,y_t,x_r,y_b\right)$ and confidence score $s$. The aim of tracking is to maximizes the posterior probability of data association given input detections and to look for a group of $N_t$ trajectories $\emph{T} = \left\{T_1,...,T_{N_t} \right\}$. By assuming mutual independence of the tracks and conditional independence of the detections given the tracks, the optimization problem is:\\ 
\begin{equation}
\begin{split}
        \emph{T}^* = \underset{\emph{T}}{\arg \max}  \prod_{j} p \left( T_j \right) \cdot \prod_{i} p \left( d_i | \emph{T} \right)
\end{split}
\end{equation}
where $p \left( d_i | \emph{T} \right)$ represents the likelihood of detection $d_i$ being associated with track $T$, while $p \left( T_j \right)$ indicates the probability of selecting a sequence of detections for track $T_j$.
By introducing the disjoint path constraints, this formulation can be translated into a cost-flow network denoted as $G(V, E, C)$, where source node is $s$ and sink node is $t$ \cite{lizhangGlobalDataAssociation2008}. Discovering the optimal association hypothesis $\emph{T}^*$ is analogous to directing the flow $f^*$ from source $s$ to sink $t$, aiming to minimizes the overall cost. The constraint of flow conservation ensures that trajectories do not overlap.
Therefore, the above question can be formulated as follows:
\begin{equation}
    \begin{split}
    & f^* =   min_f \sum_{(i,j)\in E} C(i,j) f_{ij} \\
    & s.t. f_{ij} \in \left\{0,1\right\}, \forall(i,j) \in E \\
    & \sum _{i} f_{ij} = \sum_{k} f_{jk},\forall j\in V \backslash \left\{s,t\right\}\\
    \end{split}
\end{equation}

\noindent 
where $f_{i,j}\in\left\{0,1\right\}$ is a binary vector in the flow graph which shows whether there is a flow between $i$ and $j$, each element in $C$ is a real arc cost. In this paper, we adopt an efficient muSSP algorithm \cite{wang_mussp_2019} to solve this problem. Specifically,  each detection is shown as a pair of nodes, consisting of a pre-node and a post-node. An observation edge connects these nodes, where the pre-node is appended to the source node $s$ via an enter edge, and the post-node is connected to the sink node $t$ via an exit edge. For the detection with the number $i$, the numbering rules are as follows: the pre-node is $2i-1$, and the post-node is $2i$. A transition edge emerges from the previous pre-node for every transition between detections and terminates on the post-node in the succeeding frame. Fig. \ref{fig:ssp} presents an instance of the cost-flow network.

\begin{figure}
\centering
\centerline{\includegraphics[width=0.75\columnwidth]{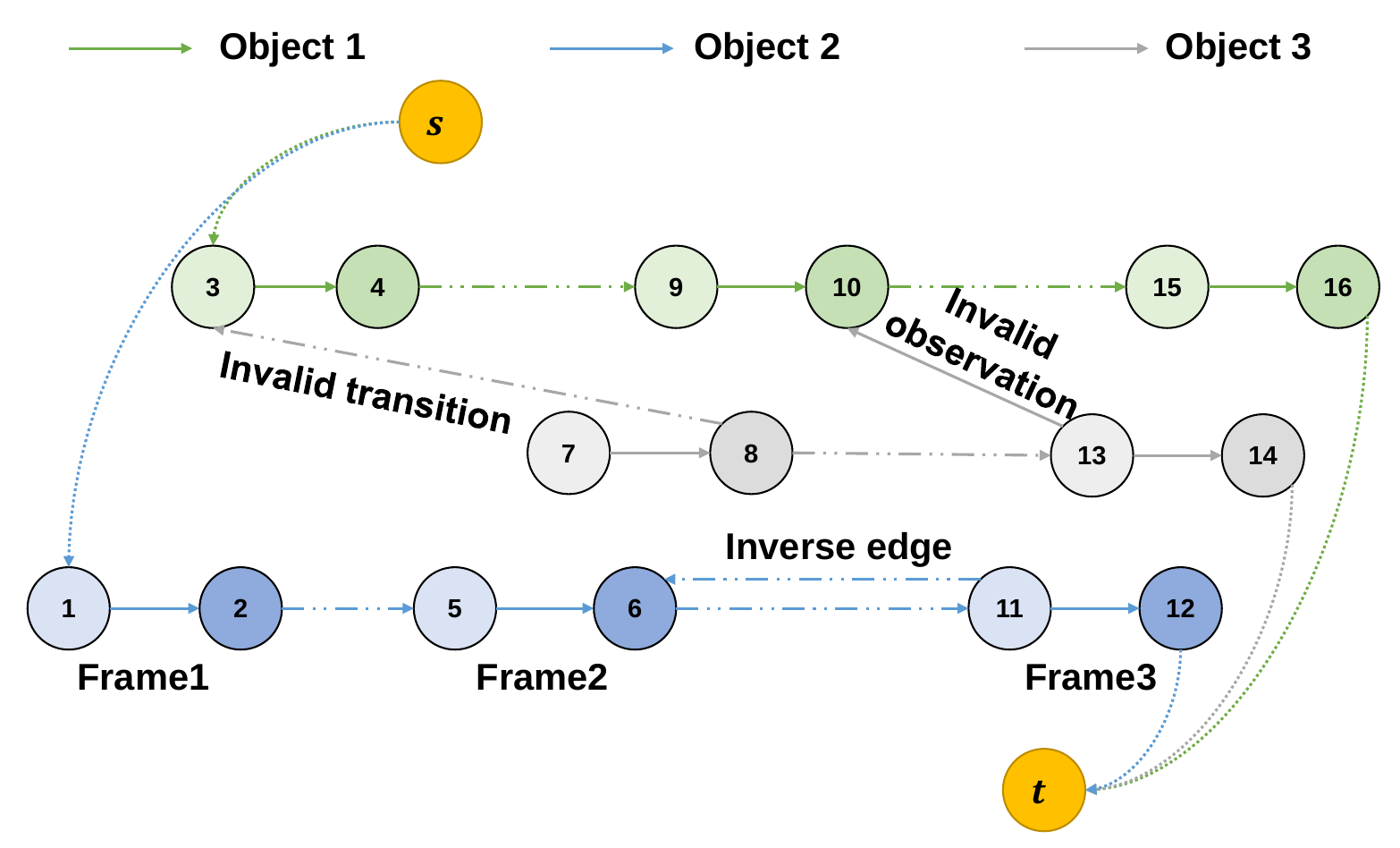}}
\caption{An example of invalid edges in min-cost flow network: invalid transition, inverse edge, and invalid observation.}
\label{fig:invalid}
\vspace{-10pt}
\end{figure}

\subsection{Data Preparation}
We first divide each video into overlapping temporal window $W$ of size $T_w$. As shown in Fig. \ref{fig:fig1}, we collect raw pictures and then detect all the objects in a window and extract the appearance feature. Detections in a window are divided into two subsets by confidence threshold $\tau$.
Detections boxes with scores surpassing $\tau$ are allocated to $D_{high}$, while those with lower scores are assigned to $D_{low}$, where $D_{high}$ and $D_{low}$ are the sets of high-confidence detections and low-confidence detections respectively.

\subsection{Tracklet Generation and Refinement}
\noindent 
\textbf{Stage1: Tracklet generation}.
This part is shown in the middle of Fig. \ref{fig:fig1} as Coarse Candidate Tracklet Generation (CCTG), we use detections in $D_{high}$ as nodes to construct a Graph $G_1(V_1, E_1, C_1)$.
For computation convenience, the nodes in frame $t-2$ can only be linked to nodes in frames $t-1$ and $t$.
Since network-flow based methods are very sensitive to false positives, CCTG only uses high-confidence detections to generate a relatively high-quality set of candidate tracklets $T_{candidate}$.\\ 
\textbf{Stage2: Tracklet Refinement}.
When long-term occlusion occurs in crowded scenes, the generated tracklets may intersect. So we use an intersection mask to locate these tracklets. The intuition behind this is that 
even though the cost is minimized, the orders of pre-nodes and post-nodes are incorrect. All trajectory segments with errors will be added to $T_{inaccurate}$. The rest will be added to $T_{accurate}$. $T_{candidate} = T_{inaccurate} \cup T_{accurate}$.

We observe that errors in $T_{candidate}$ can be divided into three categories: invalid transition, invalid observation and inverse edge. An example of the errors is shown in Fig. \ref{fig:invalid}. 
The transition edge that starts from one post-node of the later frame and reaches one pre-node of the earlier frame is considered an invalid transition edge because the graph is a directed graph.
The observation edge that violates the numerical relationship between the pre-node and the post-node of detection is considered an invalid observation.
In addition, inverse edges represent a reversal of the edge direction.
This pseudo-code for determining the intersection mask is shown in Algorithm~\ref{algo_index}.
Specifically, we traverse $T_{candidate}$ and find tracklets with the errors mentioned above. After masking tracklets in the candidate graph results, the high-confidence detection nodes obtained by CCTG are divided into two sets $D_{accurate}$ and $D_{inaccurate}$. Nodes in the $T_{inaccurate}$ trajectories will be placed in $D_{inaccurate}$, while nodes in $T_{accurate}$ will be placed in $D_{accurate}$. We then construct a second graph $G_2(V_2, E_2, C_2)$ by incorporating $D_{low}$ and $D_{inaccurate}$ to conduct the second stage network-flow association.

The observations in $D_{low}$ which are not utilized in the first stage contribute to correcting the trajectories of observations in $D_{inaccurate}$. Our approach employs high-confidence detections in $D_{inaccurate}$ as landmarks to guide the association of low-confidence detections in $D_{low}$, thereby alleviating the issue of false positives encountered in RTC.

In order to more comprehensively identify trajectory errors, the intersection mask sets all trajectories with errors as inaccurate while the validation mask is designed to identify more reliable trajectories in RTC. Based on the properties of min-cost flow, the trajectories generated earlier are more reliable, so the validation mask only marks the later intersecting trajectories as invalid.



\begin{algorithm}[h]
\SetKwInOut{Input}{input}
\SetKwInOut{Output}{output}
\Input{$T_{candidate} =\{T_1,T_2,\ldots ,T_N\}$: the candidate tracklets obtained by CCTG.\\
$N$: the number of tracklets in $T_{candidate}$.}
\Output{
$T_{inaccurate}$: tracklets with errors.
}
\For{$trackid1 \leftarrow 1$ \KwTo $N$}
{
    \For{edge in $T_{candidate}$[trackid1]}    
    {
        \For{$trackid2 \leftarrow 1 $ \KwTo $N$ \textbf{and} $trackid2 \neq trackid1$}
        {
            {\label{forins}
          \If{(edge in $T_{candidate}$[trackid2]) or (inverse edge in $T_{candidate}$[trackid2]) or (edge is invalid edge)}{
            Add ($T_{candidate}$[trackid1], $T_{candidate}$[trackid2]) to $T_{inaccurate}$. }
            }
        }
    }
}
\caption{Tracklets Intersection Mask}\label{algo_index}
\end{algorithm}
\subsection{Tracklet Association}
\noindent 
\textbf{Tracklet Cost Matrix}.
To compute the similarity of two tracklets $T^m$ and $T^n$, we need a robust tracklet feature representation method. Motivated by \cite{maggiolinoDeepOCSORTMultiPedestrian2023}, we use dynamic appearance according to the detection confidence score. Specifically, we adjust the $\alpha$ in the standard exponential moving average (EMA) based on the detector's confidence. The proposed method is:
\begin{equation} 
    e_t = \alpha_t e_{t-1} + (1-\alpha_t) e^{det}
\end{equation}
The appearance embedding of detection in the tracklet is represented by $e^{det}$, and the tracklet's appearance embedding at time $t$ is represented by $e_t$. This adaptive $\alpha_t$ enables appearance information to be selectively integrated into a tracklet during high-quality scenarios. $\alpha_t$ is calculated as follows:
\begin{equation}
    \alpha_t = \alpha_0 + (1-\alpha_0)(1-\frac{s-\sigma}{1-\sigma})
\end{equation}
Here, $s$ represents the detection confidence score and $\sigma$ represents the detection confidence threshold used to exclude noisy detections. This is necessary since the embedding of low confidence detections is always unreliable. $\alpha_0 = 0.95$ is a momentum term in the above equation. 
The similarity of appearance is determined by the cosine similarity of $e_t^p$ and $e_t^q$. Thus, we can obtain the appearance cost matrix $C^a$.

In the current window, we use regression to predict the object locations in the next window and to calculate the motion similarity of the tracklets. The motion cost is denoted by $C^m$.

To combine the motion and appearance information of the tracklet. The fused tracklet cost matrix $C^f$ can be expressed as follows:
\begin{equation}
    \bar{C}^a_{i,j} = \left\{
    \begin{aligned}
     & \gamma \cdot C^a_{i,j},(C^a_{i,j} < \theta_{emb}) \wedge (C^m_{i,j} < \theta_{iou})  \\
     & 1, otherwise  \\
    \end{aligned}\right.
\end{equation}
\begin{equation}
    C_{i,j}^f = min\left\{C^m_{i,j},\bar{C}^a_{i,j} \right\}
\end{equation}
Where $\gamma = 0.5$, $C^m_{i,j}$ and $C^a_{i,j}$ respectively represent the motion cost and appearance cost 
between $i$-th tracklet in current window and $j$-th tracklet in the next window. We reject low cosine similarity or far away candidates to get new appearance cost $\bar{C}^a_{i,j}$. $\theta_{iou}$ is iou threshold to reject unlikely pairs of tracklets. $\theta_{emb}$ is the appearance threshold to distinguish different object embeddings.\\
\textbf{Hierarchical Association}.
As a single matching does not reflect the order of priority within tracklet association, we split the tracklets into high and low confidence sets during the tracklet association phase. We initially associate all tracklets in the current window to high-confidence tracklets in the next window through a matching threshold of $\theta^{match}_1$. Next, we use unmatched tracklets in the first association to match the low-confidence tracklets in the next window, utilizing a different matching threshold of $\theta^{match}_2$. 

\section{Experiments and Results}
\subsection{Implementation}
Experiments are performed in three representative MOT datasets: MOT16 \cite{milanMOT16BenchmarkMultiObject2016}, MOT17 and MOT20 \cite{dendorferMOT20BenchmarkMulti2020}. These datasets presents chanllenges due to frequent occlusions, diverse camera viewpoints, human poses variations and complex backgrounds. Notably, both The MOT16 and MOT17 datasets consist of videos that capture camera movement. In alignment with prior research in MOT16 and MOT17, MOT20 exhibits a notably higher density of pedestrians within the scenes, longer sequences, and more frequent occurrences of occlusions.

To evaluate our proposed method, we use the primary evaluation metric HOTA \cite{luitenHOTAHigherOrder2021}, 
MOTA \cite{bernardinEvaluatingMultipleObject2008}, MT (mostly tracked targets), ML (mostly lost targets), IDS (ID switches), and IDF1 \cite{ristani2016performance}.
While MOTA emphasizes detection performance, IDF1 evaluates the preservation of identities, focusing on association performance. HOTA offers a more comprehensive measure, striking a balance between accurate detection, association, and localization.
For equitable comparisons, we employ YOLOX trained by \cite{zhang_bytetrack_2022} for MOT17 and MOT20, along with the FastReID's \cite{He2020FastReIDAP} SBS-50 model trained by \cite{aharonBoTSORTRobustAssociations2022} as feature extractor. The detection score threshold $\tau = 0.6$. In the tracklet association step the default values of $\theta^{match}_1$ and $\theta^{match}_2$ are 0.7 and 0.5. Unmatched trajectories are retained for 30 frames in case they reappear. In scenarios involving significant camera motion, the bounding boxes from previous frame may fail to enclose the tracked object in the next frame. To address this, we implement a camera motion compensation (CMC) \cite{evangelidis2008parametric}. We align the frames by image registration to reveal the background motion.
\subsection{Comparison with the state-of-the-art}
We compare our approach with the state-of-the-art trackers in Table \ref {tab:tab1} using private detections. We present results on the testing sets of the MOT16, MOT17, and MOT20 datasets respectively. The entirety of the outcomes is sourced directly from the official evaluation server of the MOT challenge. Our approach outperforms all compared trackers on MOT16. The results show that our method has impressive tracking performance in scenes with camera motion. Furthermore, our proposed method shows remarkable scalability in crowded scenes. Despite trailing Bytetrack \cite{zhang_bytetrack_2022} one point behind on the HOTA metric, our approach demonstrates superior tracking performance in terms of MT and ML metrics. 
\begin{table}[!t]
    \renewcommand{\arraystretch}{1.3}
    \setlength{\tabcolsep}{3pt}
    \caption{Comparison of the state-of-the-art trackers using private detections on the test set of MOT16, MOT17, MOT20. \textbf{Boldface} indicates the best results.}
    \resizebox{\columnwidth}{!}
    {
    \begin{tabular}{cccccccccc}
    \toprule
        \textbf{Dataset} & \textbf{Method} & \textbf{MOTA} $\uparrow$ & \textbf{IDF1} $\uparrow$ & \textbf{HOTA}$\uparrow$ & \textbf{MT} $\uparrow$ &
        \textbf{ML} $\downarrow$ & \textbf{AssA} $\uparrow$ & \textbf{DetA} $\uparrow$ &\textbf{LocA} $\uparrow$ \\ 
        \hline
        \multirow{5}{*}{MOT16}
        & TubeTK \cite{pangTubeTKAdoptingTubes2020} & 66.9 & 62.2 & 50.8 & 296 & 122 & 47.3 & 55.0 & 81.2 \\
        & QuasiDense \cite{pangQuasiDenseSimilarityLearning2021} & 69.8 & 67.1 & 54.5 & 316 & 150 & 52.8 & 56.6 & 81.7\\
        & TraDeS \cite{wuTrackDetectSegment2021} & 70.1 & 64.7 & 53.2 & 283 & 152 & 50.9 & 56.2 & 81.7\\
        & ReMOT \cite{yangReMOTModelagnosticRefinement2021a}& 76.9 & \textbf{73.2} & 60.1 & 390 & 94  & 57.8 & 62.9 & 82.8\\
        & TSMCF & \textbf{78.4} & 72.9 & \textbf{61}   & \textbf{407} & \textbf{92}  & \textbf{58.4} & \textbf{64}   & \textbf{83.1}\\
        \hline
        \multirow{6}{*}{MOT17}
        & CenterTrack \cite{zhouTrackingObjectsPoints2020} & 67.8 & 64.7 & 52.2 & 816 & 579 & 51.0 & 53.8 & 81.5\\
        & FairMOT \cite{zhangFairMOTFairnessDetection2021}     & 73.7 & 72.3 & 59.3 & 1017 & 408  & 58.0 & 60.9  & 83.6 \\
        & TrackFormer \cite{meinhardtTrackFormerMultiObjectTracking} & 74.1 & 68.0 & 57.3 & 1113 & 246  & 54.1 & 60.9 & 82.8\\
        & TransTrack \cite{sunTransTrackMultipleObject2021}  & 75.2 & 63.5 & 54.1 & 1302 & \textbf{240}  & 47.9 & 61.6 & 82.6 \\ 
        & MAA \cite{stadlerModellingAmbiguousAssignments2022}& \textbf{79.4} & \textbf{75.9} & \textbf{62.0} & \textbf{1356} & 282  & \textbf{60.2} & \textbf{64.2} & 82.8\\
        & TSMCF & 79.2 & 72.9 & 60.7 & 1236 & 309 & 57.9 & 64   & \textbf{83.2} \\
        \hline
        \multirow{8}{*}{MOT20}
         & FairMOT \cite{zhangFairMOTFairnessDetection2021}    & 61.8 & 67.3 & 54.6 & 855 & 94  & 54.7 & 54.7  & 81.1 \\
         & TransTrack \cite{sunTransTrackMultipleObject2021} & 65.0 & 59.4 & 48.9 & 622 & 167 & 45.2 & 53.3  & 82.8 \\
         & MAA \cite{stadlerModellingAmbiguousAssignments2022}        & 73.9 & 71.2 & 57.3 & 741 & 153 & 55.1 & 59.7  & 82.7 \\
         & QD-Track \cite{pangQuasiDenseSimilarityLearning2021}   & 74.7 & 73.8 & 60.0 & 797 & 162 & 58.9 & 61.4 & 83.8  \\
         & OC-SORT \cite{caoObservationCentricSORTRethinking2022}    & 75.7 & \textbf{76.3} & 62.4 & 813 & 160 & \textbf{62.5} & 62.4 & \textbf{84.2}\\
       & ReMOT \cite{yangReMOTModelagnosticRefinement2021a}      & 77.4 & 73.1 & 61.2 & 846 & 123 & 58.7 & \textbf{63.9} & 84   \\
       & Bytetrack \cite{zhang_bytetrack_2022}  & \textbf{77.8} & 75.2 & \textbf{61.3} & 859 & 118 & 59.6 & 63.4 & 83.6 \\
       & TSMCF        & 76.4 & 71.6 & 60.3 & \textbf{886} & \textbf{91}  & 57.5 & 63.6 & 83.8  \\
    \bottomrule
    \end{tabular}}
    \label{tab:tab1}
\end{table}

\subsection{Ablation studies}

\begin{figure}[!t]
	\centering
	\footnotesize
	\renewcommand{\tabcolsep}{1.3pt} 
	\renewcommand{\arraystretch}{1.3} 
	\begin{tabular}{c cc}
        \includegraphics[width = 0.5 \columnwidth]{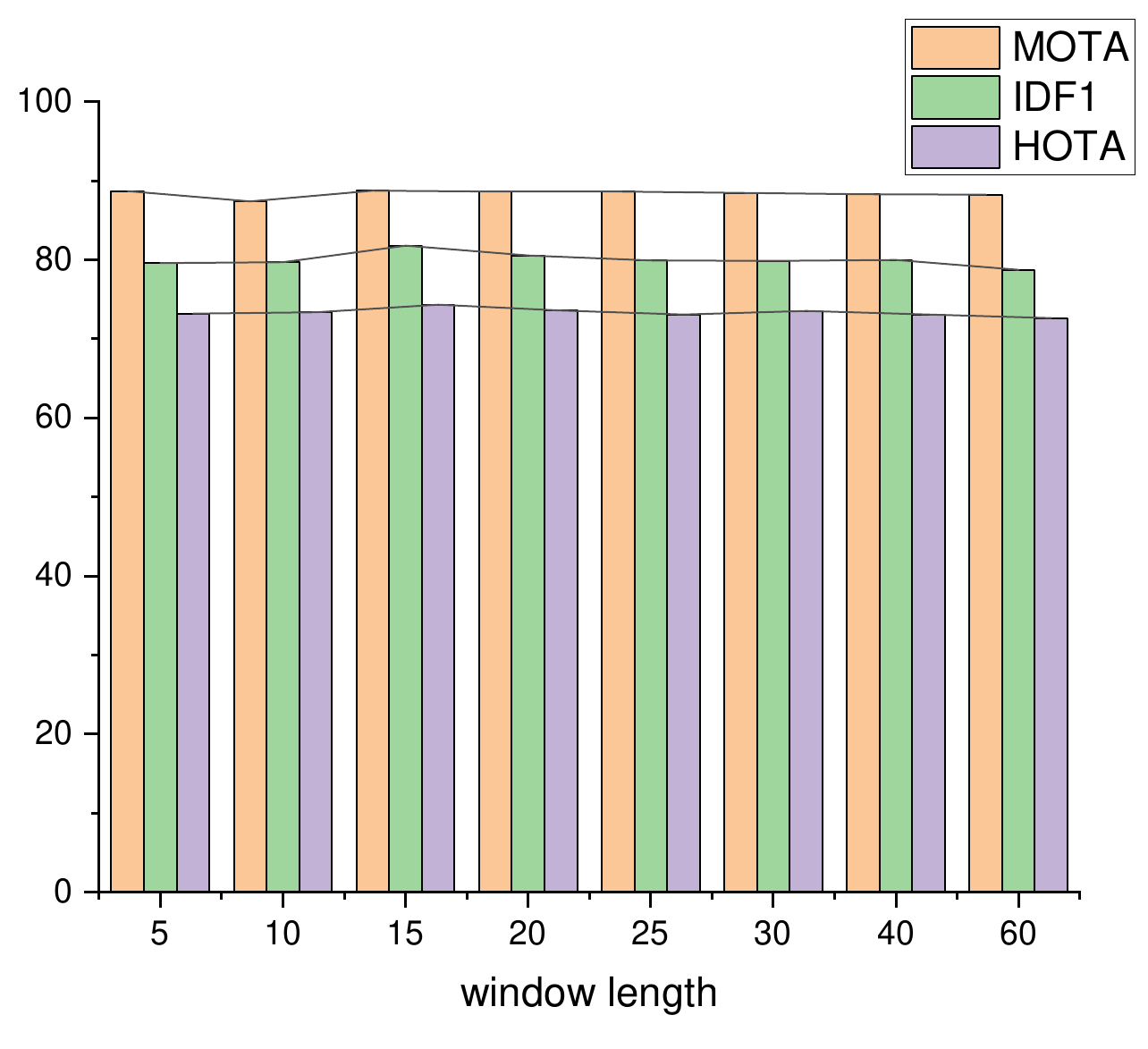} & &
        \includegraphics[width = 0.5 \columnwidth]{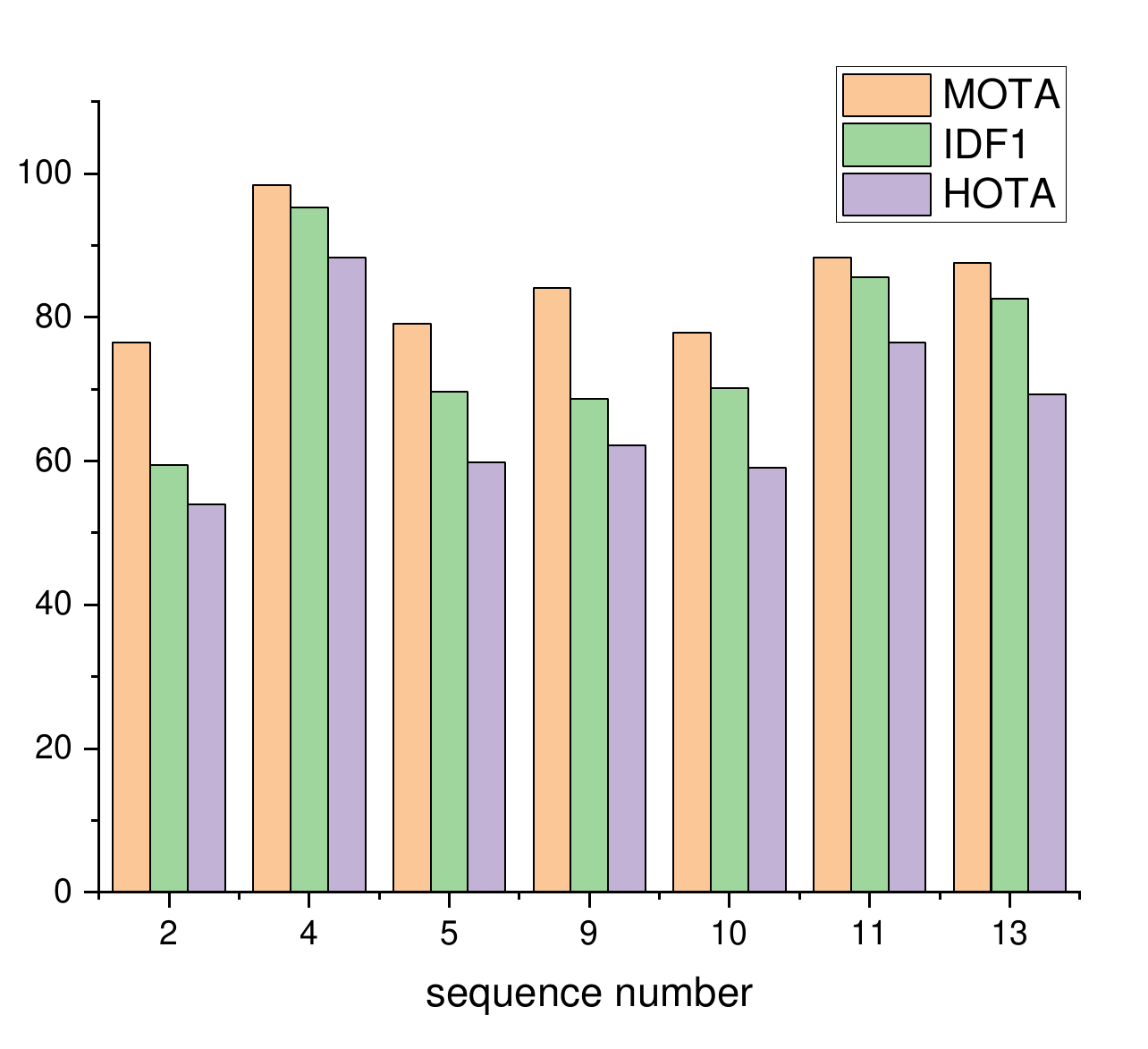} \\
        (a)  & & (b) 
    \end{tabular}
   \caption{(a) Tracking performance over different
window sizes on the MOT16 train set.  (b) Fixed window size $T_w = 15$ for different sequences on MOT16 train set.}
	\label{fig:window_size}
	\label{fig:compression}
\end{figure}

\noindent 
\textbf{Design Choices for Window Size}. 
In Fig.  \ref{fig:window_size} we emphasize the impact on performance when selecting varying sizes for the temporal window $T_w$. Our experiment is conducted on MOT16. The temporal window $T_w$ ranges from 5 to 60.
The variation of the three metrics is very small, which shows our proposed tracker is very robust to variations of $T_w$.\\
\noindent 
\textbf{Analysis of the Individual Components}.
The main objective of our ablation study is to confirm the contribution of each component. We take the single-stage min-cost flow results as a baseline and the tracklet association method only considers the adjacent frames rather than global tracklet information. Baseline1 only uses $D_{high}$ while baseline2 uses both $D_{high}$ and $D_{low}$.
We design five groups of comparative experiments, including baseline1, baseline2, baseline1 with camera motion compensation (denoted as baseline1 + C), baseline1 with proposed tracklet association and camera motion compensation (denoted by baseline1 + C + T), and the complete two-stage tracking method (denoted as baseline1 + C + T + G). Experimental results are listed in Table \ref{tab:tab2}. The significant performance difference between baseline1 and baseline2 suggests that the problem of association in the min-cost flow cannot be solely resolved by lowering the detection threshold. Adding camera motion compensation can partially enhance the tracking performance. Using the proposed tracklet association method can effectively reduce the IDS. That is because the presented tracklet representation method can incorporate object motion information and have a more robust appearance representation.
Finally, adding the correction and second graph construction proposed in this paper can greatly improve tracking performance, which shows effectiveness. When compared to baseline 2, it is more favourable to utilise $D_{low}$ in the second stage, rather than in the first stage.


\begin{table}[htbp]
    \renewcommand{\arraystretch}{1.3}
    \caption{Ablation studies on MOT16 train set. \textbf{Boldface} indicates the best results.}
    \centering
    \resizebox{\linewidth}{!}
    {
    \begin{tabular}{ccccccc}
    \toprule
        \textbf{Methods} & \textbf{HOTA} $\uparrow$ & \textbf{MOTA} $\uparrow$ & \textbf{IDF1}$\uparrow$ & \textbf{MT} $\uparrow$ &
        \textbf{ML} $\downarrow$ &\textbf{IDS} 
        $\downarrow$ \\
        \hline
        baseline1 & 69.996 & 86.035 & 75.442 & 355 & 14 & 1060 \\
        baseline2 & 64.302 & 72.792 & 67.292 & \textbf{474} & \textbf{7}  & 1750 \\
        \hline
        baseline1 + C & 70.105 & 86.177 & 75.317 & 354 & 15 & 1007 \\
        baseline1 + C + T & 70.236 & 86.204 & 75.745 & 353 & 15 & 975 \\
        baseline1 + C + T + G & \textbf{73.356} & \textbf{89.067} & 
        \textbf{79.674} & 424 & 12 & \textbf{533}\\
    \bottomrule
    \end{tabular}
    }
    \label{tab:tab2}
\end{table}
\section{Conclusion}
This paper proposes a two-stage minimum network flow correction tracker named TSMCF that can accurately locate the errors in MCF, and the two-stage method based on the intersection judgment realizes the association correction. It is not only simple but also greatly improves the effectiveness of the original algorithm, and finally achieves good results on multiple benchmarks of MOTchallenge.

\bibliographystyle{IEEEtran}
\bibliography{ref}{}

\end{document}